\documentclass[letterpaper]{article} % DO NOT CHANGE THIS
\usepackage{aaai2026}  % DO NOT CHANGE THIS
\usepackage{times}  % DO NOT CHANGE THIS
\usepackage{helvet}  % DO NOT CHANGE THIS
\usepackage{courier}  % DO NOT CHANGE THIS
\usepackage[hyphens]{url}  % DO NOT CHANGE THIS
\usepackage{graphicx} % DO NOT CHANGE THIS
\usepackage{natbib}  % DO NOT CHANGE THIS AND DO NOT ADD ANY OPTIONS TO IT
\usepackage{caption} % DO NOT CHANGE THIS AND DO NOT ADD ANY OPTIONS TO IT
\usepackage{algorithm}
\usepackage{algorithmic}
\usepackage{amssymb}
\usepackage{amsmath}
\usepackage{booktabs}
\usepackage{multirow}
\usepackage{newfloat}
\usepackage{listings}
\DeclareCaptionStyle{ruled}{labelfont=normalfont,labelsep=colon,strut=off} % DO NOT CHANGE THIS
\floatstyle{ruled}
\newfloat{listing}{tb}{lst}{}
\floatname{listing}{Listing}
\title{MUG: Meta-path-aware Universal Heterogeneous Graph Pre-Training}

\author{
    Lianze Shan\textsuperscript{\rm 1}\equalcontrib, 
    Jitao Zhao\textsuperscript{\rm 1}\equalcontrib, 
    Dongxiao He\textsuperscript{\rm 1}\thanks{Corresponding Author.}, 
    Yongqi Huang\textsuperscript{\rm 1}, 
    Zhiyong Feng\textsuperscript{\rm 1}, 
    Weixiong Zhang\textsuperscript{\rm 2}
}
\affiliations{
    \textsuperscript{\rm 1}College of Intelligence and Computing, Tianjin University, Tianjin, China\\
    \textsuperscript{\rm 2}Departments of Health Technology \& Informatics; Data Science \& Artificial Intelligence; and Computing,\\
    The Hong Kong Polytechnic University, Kowloon, Hong Kong\\
    \{shanlz2119, zjtao, hedongxiao, yqhuang, zyfeng\}@tju.edu.cn,
    weixiong.zhang@polyu.edu.hk

}

\begin{document}

\maketitle

\begin{abstract}
Universal graph pre-training has emerged as a key paradigm in graph representation learning, offering a promising way to train encoders to learn transferable representations from unlabeled graphs and to effectively generalize across a wide range of downstream tasks. However, recent explorations in universal graph pre-training primarily focus on homogeneous graphs and it remains unexplored for heterogeneous graphs, which exhibit greater structural and semantic complexity. 
This heterogeneity makes it highly challenging to train a universal encoder for diverse heterogeneous graphs: (i) the diverse types with dataset-specific semantics hinder the construction of a unified representation space;  (ii) the number and semantics of meta-paths vary across datasets, making encoding and aggregation patterns learned from one dataset difficult to apply to others. 
To address these challenges, we propose a novel Meta-path-aware Universal heterogeneous Graph pre-training (MUG) approach. 
Specifically, for challenge (i), MUG introduces a input unification module that integrates information from multiple node and relation types within each heterogeneous graph into a unified representation.
This representation is then projected into a shared space by a dimension-aware encoder, enabling alignment across graphs with diverse schemas.
Furthermore, for challenge (ii), MUG trains a shared encoder to capture consistent structural patterns across diverse meta-path views rather than relying on dataset-specific aggregation strategies, while a global objective encourages discriminability and reduces dataset-specific biases.
Extensive experiments demonstrate the effectiveness of MUG on some real datasets.
\end{abstract}

\begin{links}
    \link{Code}{https://github.com/slz1024/MUG}
\end{links}

\section{Introduction}
\label{sec:introduction}
Universal Graph Pre-training (UGP), which aims to build strong, highly generalizable models applicable to a wide range of downstream applications, has attracted considerable interest lately \cite{gcc,graphcontrol}. 
Unlike traditional pre-training approaches that are tailored to specific tasks or graphs \cite{GOUDA, NeCo, ga-ggd, HEATS}, UGP aims to learn graph representations that capture shared structural and semantic patterns, facilitating effective transfer of learned representations across tasks and datasets without extensive re-training. 
The transition from building individual models for specific tasks to designing reusable encoders has advanced the development of graph foundation models \cite{gft,graphclip,Uniprompt}.

Despite these advances, most existing UGP methods \cite{FUG, MDGPT,samgpt} focus on homogeneous graphs that involve a single node type and relatively simple, fixed relations.
In fact, graphs in real world are typically heterogeneous \cite{HAN}, such as user–item interactions for recommendation \cite{LightGCN}, knowledge graphs\cite{heter_knowledge_graph}, and social networks \cite{socialnetwork}, which contain multiple node and edge types and complex semantics. 
This heterogeneity carries richer schema-specific signals and relational patterns that cannot be effectively modeled by homogeneous graph approaches.
Outside the UGP paradigm, in the classical graph heterogeneous learning field, many non-universal Heterogeneous Graph Representation Learning methods (HGRLs) have been proposed to capture rich semantics in heterogeneous graphs \cite{hgrl_survey}. 
These HGRLs, whether supervised or self-supervised, have effectively modeled semantic complexity and demonstrated strong performance \cite{HAN,HeCo,HGMAE,hero}.

However, the pursuit of a universal modeling framework has not aligned well with the ubiquitous heterogeneous graphs. 
To the best of our knowledge, no method has been developed so far for universal heterogeneous graph pre-training. 
This significant gap is mainly attributed to two factors.
The first is that existing UGP methods assume each dataset contains a fixed and consistent set of node and relation types, enabling unified input spaces and reusable encoders.
Recent works, such as FUG \cite{FUG} and SAMGPT \cite{samgpt}, can achieve attribute and structure transfer across graphs, respectively. 
Nevertheless, both of them rely on each dataset having a fixed set of entity and relation types. 
In heterogeneous graphs, however, the entity and relation types within each graph vary significantly, making these methods ineffective. 
The second factor is the limited generalizability of traditional HGRLs. These methods are typically designed with dataset-specific assumptions, including type-specific encoders, relation-aware aggregation schemes, and meta-path semantics tailored to particular graph schemas. 
As a result, they are tightly coupled to the training data and fail to generalize to unseen heterogeneous graphs

The above observation naturally raises a question: What are the key issues that must be addressed to enable effective universal pre-training on heterogeneous graphs?
We identify two primary challenges. 
\textbf{First,} each heterogeneous graph inherently contains multiple types of nodes and relations, making it difficult to unify. 
Different heterogeneous graphs may contain completely different node types, relation types, and node attributes.
For example, the ACM dataset \cite{ACM} involves papers, authors, and subjects, connected via paper–author and paper–subject relations, while the Freebase dataset \cite{freebase} includes movies, actors, directors, and writers, with relations such as movie–actor, movie–director, and movie–writer. 
These differences are not only in the number of types but also in their semantics and relational patterns. 
As a result, it is difficult to design a universal model to directly encode multiple heterogeneous graphs without first addressing the mismatch in node types, relation types, and node attributes. 
\textbf{Second,} the information learned from heterogeneous graphs is hard to transfer. 
This stems from two core components of graph representation learning: encoding and aggregation.
Existing HGRLs typically require the design of distinct encoding functions to encode different views of a node, and rely on task-specific or data-specific objective functions to learn aggregation function tailored to the given graph. 
As a result, the learned representations and aggregation patterns (such as attention weights or transformation matrices), tend to be tightly coupled with the characteristics of the training dataset, making it difficult to generalize or transfer effectively to unseen heterogeneous graphs.

\begin{figure}[H]
\includegraphics[width=1.0\linewidth]{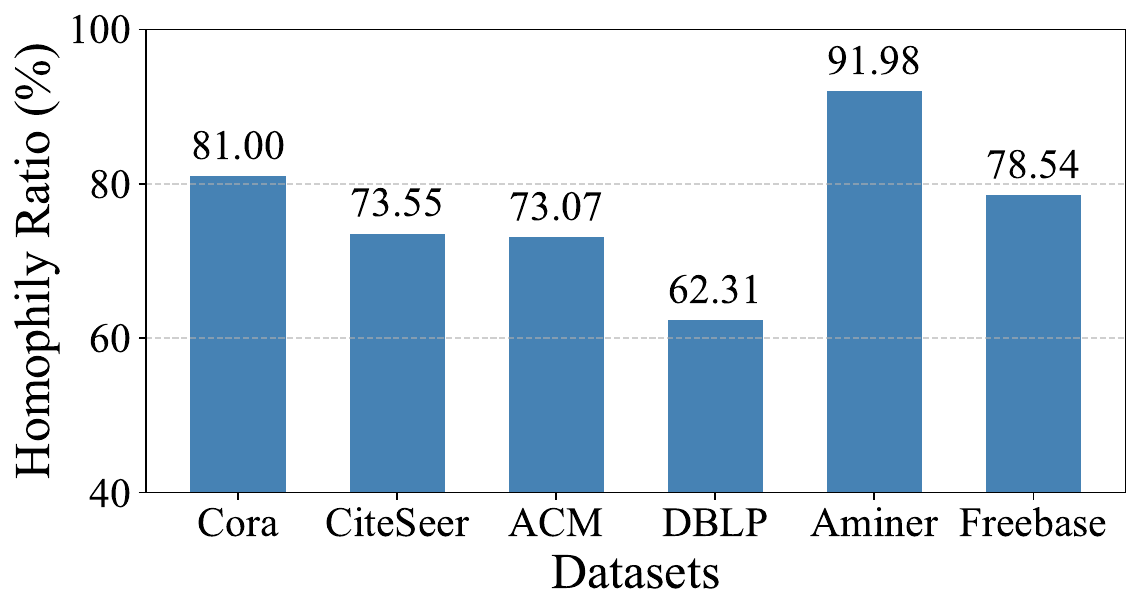}
    \caption{Homophily ratios on six datasets. Cora and CiteSeer  \cite{Cora}  are homogeneous graphs, while ACM \cite{ACM}, DBLP \cite{magnn}, AMiner \cite{Aminer} and Freebase \cite{freebase} are heterogeneous. For heterogeneous graphs, we compute the homophily ratio for each meta-path-based adjacency matrix and report the average. }
    \label{fig:motivation}
\end{figure}

To address these challenges, we propose a novel \textbf{M}eta-path-aware \textbf{U}niversal heterogeneous \textbf{G}raph pre-training approach (MUG).
To address the first challenge, we propose a input unification module, which aims to encode heterogeneous types and attributes into a unified input representation and builds a shared representation space for different heterogeneous graphs.
Specifically, we generate a new embedding for each node by encoding its contextual structural information.
This enables the model to capture high-order semantic relationships across heterogeneous graphs, such as co-authorship in ACM and co-acting in Freebase. 
We then get a unified input representation by concatenating this contextual embedding with the original node attributes, and align the representation space by introducing a dimension-aware encoder.
For the second challenge, we start from a key observation: in homogeneous graphs, transferable encoding and aggregation strategies have been successfully developed based on homophily assumption. 
To explore whether a similar property exists in heterogeneous graphs, we conduct a motivation experiment as shown in Figure \ref{fig:motivation}. 
It can be observed that the average homophophily ratio across meta-paths is comparable to that of the homogeneous graphs. 
This suggests that a shared encoder can be used across meta-path views.  
And we can design an objective function , which is based on the homophily assumption, to guide the encoder to capture the common connectivity patterns of neighboring nodes under different views.
Furthermore, we introduce a global scattering objective function during aggregation. 
It encourages globally discriminative embeddings to guide aggregation instead of relying on dataset-specific prior knowledge, reducing overfitting to training graphs. 
Our contributions can be summarized as follows:
\begin{itemize}
    \item To the best of our knowledge, this is the first work to explore LLM-free universal pre-training approach on heterogeneous graphs. It enables a model pre-trained on one heterogeneous graph and then to be generalized to unseen datasets without any retraining, greatly extending the applicability of graph foundation models to heterogeneous application domains.
    \item We propose MUG, a novel universal heterogeneous graph pre-training approach. MUG constructs a unified representation space and learns transferable information through universal encoding and aggregation .
    \item We conduct extensive experiments on four datasets under cross-domain and few-shot settings. MUG consistently outperforms all baseline methods, demonstrating its strong generalization ability and effectiveness across diverse scenarios.
\end{itemize}

\section{Related Work}
\textbf{Heterogeneous Graph Representation Learning.} This paradigm aims to encode the rich structural and semantic heterogeneity of graphs into low-dimensional embeddings that are suitable for downstream tasks.
Early works in HGRLs focus on supervised or semi-supervised heterogeneous graph neural networks, which encode nodes and relations through meta-path-based or type-specific parameters.
For example, HAN \cite{HAN} employs a hierarchical attention mechanism to model the importance of different neighbors and their corresponding semantic relations. 
MAGNN \cite{magnn} captures semantics by jointly modeling node attributes and intermediate nodes along meta-paths, followed by aggregation across multiple meta-paths.
HGT \cite{hgt} extends the transformer architecture to heterogeneous graphs by utilizing type-specific parameters and attention, thereby enabling effective large-scale modeling. 
Despite their effectiveness, these methods rely heavily on labeled data, which limits their application to problems with data with a limited number of labels.
To overcome this limitation, heterogeneous graph self-supervised learning has attracted increasing interest.
It exploits the intrinsic structural and semantic information of heterogeneous graphs and designs pretext tasks to train heterogeneous graph neural network encoders without any labels.
For example, HeCo \cite{HeCo} constructs dual views based on the network schema and meta-paths and performs cross-view contrastive learning to capture both local and high-order semantics. 
HGCL \cite{HGCL} further improves contrastive learning by introducing attribute and topology views, aligning them through a reciprocal contrastive mechanism to improve robustness.
HGMAE \cite{HGMAE} employs a masked autoencoder paradigm, training the encoder to reconstruct structural, attribute, and positional information.
However, these methods trained on one dataset struggle to generalize to unseen datasets, which hinders the development of graph foundation models.

\textbf{Universal Graph Pre-training.} This paradigm has emerged as a promising direction to learn transferable representations across diverse graphs without relying on task-specific or data-specific supervision \cite{All-in-One}. 
Existing methods typically unify input spaces through structural encodings \cite{gcc,samgpt}, attribute projection \cite{FUG,MDGPT}, or attribute textualization \cite{OFA}.
While these methods reduce distribution shifts and improve generalization, they are primarily designed for homogeneous graphs with uniform node and edge types and thus struggle to capture the rich semantics and relational diversity inherent in heterogeneous graphs. 
\textbf{Notably}, HiGPT \cite{higpt} investigates cross-domain transferable representation learning for heterogeneous graphs by incorporating Large Language Models (LLM). However, its reliance on the textualization of node attributes poses challenges for generalizing to broader heterogeneous graphs with sparse or non-textual attributes. 
Our work instead focuses on developing a LLM-free universal pre-training approach for heterogeneous graphs, enabling broader applicability across diverse real-world scenarios.

\section{Preliminary}
\label{sec:preliminary}
We now introduce the notations and preliminaries adopted in the paper. 
We use script letters (e.g., $\mathcal{V}, \mathcal{E}$) to represent sets, bold uppercase letters 
(e.g., $\mathbf{H}$) to indicate matrices, bold lowercase letters (e.g., $\mathbf{h}_i$) to represent 
vectors (typically rows or columns from a matrix), and plain lowercase letters (e.g., $h$) to denote scalar values. 
All graphs, node attributes, meta-paths, and model outputs adhere to this unified notation, unless otherwise specified.

\textbf{Definition 3.1. Heterogeneous Graph.} A heterogeneous graph is defined as $\mathcal{G} = (\mathcal{V}, \mathcal{E}, \mathcal{X}, \mathcal{A}, \mathcal{R})$, where $\mathcal{V}$ is the set of nodes, $\mathcal{E}$ is the set of edges, $\mathcal{X}$ is the set of node attribute matrices, $\mathcal{A}$ and $\mathcal{R}$ are finite sets of node types and edge (relation) types respectively. 
Every node $v \in \mathcal{V}$ is associated with a type of assignment via a mapping  $\phi_v: \mathcal{V} \rightarrow \mathcal{A}$, and every edge $e \in \mathcal{E}$ is associated with a relation type $\phi_e: \mathcal{E} \rightarrow \mathcal{R}$. The graph is considered  \textit{heterogeneous} when $|\mathcal{A}| + |\mathcal{R}| > 2$, indicating multiple node or edge types.
A dataset-specific graph $\mathcal{G}^{(i)}$ follows this unified formulation but can be instantiated with different types of sets $\mathcal{A}^{(i)}$ and $\mathcal{R}^{(i)}$.

\textbf{Definition 3.2. Pretraining and Evaluation Datasets.}  We denote a heterogeneous graph pre-training dataset as 
$\mathcal{D}_{\text{train}} = \{ \mathcal{G}^{(1)}, \mathcal{G}^{(2)}, \dots, \mathcal{G}^{(t)} \}$, where each $\mathcal{G}^{(i)}$ represents an individual heterogeneous graph instance. A universal encoder with parameters $\theta$ $f_{\theta}\colon \mathcal{V} \rightarrow \mathbb{R}^k$ is trained on$\mathcal{D}_{\text{train}}$, where $k$ is the embedding dimension.
During downstream evaluation, the learned encoder is transferred without any parameter updates to a set of unseen heterogeneous graphs $\mathcal{D}_{\text{test}} = \{ \mathcal{G}^{\prime(1)}, \mathcal{G}^{\prime(2)}, \dots , \mathcal{G}^{\prime(t')}  \}$. The encoder $f_{\theta}$ is fixed, and its output representations are used directly for downstream tasks such as node classification.

\textbf{Definition 3.3. Meta-path.} A meta-path $\mathcal{P}$ is defined as a sequence of node and relation types in the form of $\mathcal{P} : A_1 \xrightarrow{R_1} A_2 \xrightarrow{R_2} \cdots \xrightarrow{R_l} A_{l+1},$ where $A_i \in \mathcal{A}$ and $R_i \in \mathcal{R}$. Each meta-path represents a composite semantic relation connecting nodes of type $A_1$ and $A_{l+1}$ through a sequence of relations. For instance, a meta-path Author-Paper-Venue reflects a two-hop semantic dependency from authors to venues.
In our setting, meta-paths serve as structural and semantic templates that guide the modeling of contextual information across graphs. 
Although different datasets may define distinct meta-paths, many of them reflect transferable structural or semantic patterns across heterogeneous graphs. 
Our model is designed to extract such patterns and apply them across heterogeneous graphs.

\section{Method}
\begin{figure*}[t]
\centering
\includegraphics[width=1\textwidth]{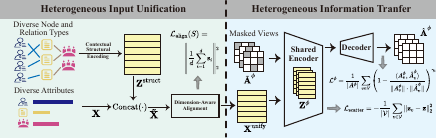} 
\caption{Overview of the MUG. Given a heterogeneous graph with diverse node types, relation types, and attributes, we first get embedding $\mathbf{Z}^\text{struct}$ by contextual structural encoding. The embedding is concatenated with the original node attributes $\mathbf{X}$ to obtain unified representation $\tilde{\mathbf{X}}$, which contains diverse type and attribute information. Then, a dimension-aware encoder is applied to align representation spaces and produce the unified input $\mathbf{X}^\text{unify}$. 
Finally, a shared GNN encoder is used to encode each masked adjacency matrix $\tilde{\mathbf{A}}^\phi$  with the unified input $\mathbf{X}^\text{unify}$, producing $\mathbf{Z}^\phi$. The model is optimized by three objectives: a dimension alignment loss $\mathcal{L}_\text{align}$, a meta-path masked reconstruction loss $\mathcal{L}^\phi$ and a global scattering regularization loss $\mathcal{L}_\text{scatter}$.
}
\label{fig:overview}
\end{figure*}

In this section, we present MUG, a novel universal pre-training approach designed for heterogeneous graphs. 
As shown in Figure \ref{fig:overview}, MUG integrates two core components: a module that unifies variety of heterogeneous input by encoding context structural information and aligning diverse heterogeneous representation spaces; and a module that enables transferable heterogeneous information including encoding and aggregation.
We discuss each component in the following subsections.

\subsection{Heterogeneous Input Unification}

\textbf{Contextual structural encoding.} As defined in preliminary, nodes and edges in each heterogeneous graph may be associated with multiple types, leading to diversity in node types (e.g., papers, authors, conferences), relation types (e.g., writes, published-in), and node attributes.
A typical heterogeneous graph neural network encodes these rich semantics by defining type-specific transformation matrices and encoders \cite{HeCo, HGMAE}. 
For a node $v$ with type $\phi_i$, a type-specific transformation matrix $\mathbf{W}_{\phi_i}^v$ is applied to project its original attributes $\mathbf{x}_v$ into a shared space:
\begin{equation}
    \mathbf{h}_i = \sigma(\mathbf{W}_{\phi_i} \mathbf{x}_v + \mathbf{b}_{\phi_i}),
\end{equation}
where the number and semantics of the parameters set $\{\mathbf{W}_{\phi_i}, \mathbf{b}_{\phi_i} \}$ are tightly coupled with the node type $\phi_i$, making them inherently dependent on the schema of the given heterogeneous graph. 
Similarly, each relation type $r$ is associated with a dedicated encoder $\text{Enc}_r(\cdot)$:
\begin{equation}
    \mathbf{z}_i^{(r)} = \text{Enc}_r\left( \left\{ h_j \mid j \in \mathcal{N}_i^{(r)} \right\} \right).
\label{eq:hgrl_encoding}
\end{equation}
However, node and relation types may vary significantly in both number and semantics across datasets, making the transformation matrices and encoders defined on one specific dataset difficult to transfer to another.

To enable unified input across heterogeneous graphs, we construct contextual  structural embeddings independent of specific node and relation types.
Instead of encoding type identities directly, we embed each node through the semantic structures formed by its interactions with other node types, captured via meta-paths. 
Formally, given a heterogeneous graph $\mathcal{G}$ and a set of meta-paths $\mathcal{P} = \{ \mathcal{P}_1, \mathcal{P}_2, \dots, \mathcal{P}_L \}$, we define the structural embedding of node $v$ as:
\begin{equation}
\mathbf{z}_v^{\text{struct}} = \mathcal{F}_{\text{context}}(v, \mathcal{G}, \mathcal{P}),
\end{equation}
where $\mathcal{F}_{\text{context}}(\cdot)$ denotes a general structural embedding function that can be instantiated by methods capable of capturing high-order semantic dependencies across different node and relation types.
Inspired by \cite{metapath2vec}, we instantiate $\mathcal{F}_{\text{context}}$ using a meta-path guided context encoder, which samples node sequences along meta-paths and optimizes node embeddings to predict their structural context.
Specifically, for each meta-path $\mathcal{P}_\ell \in \mathcal{P}$, we perform meta-path guided random walks to sample structural contexts for each node $v$:
\begin{equation}
\mathcal{C}_v^{(\ell)} = \left\{
(v_0 = v, v_1, \ldots, v_K) \;\middle|\;
\begin{aligned}
    &(v_i, v_{i+1}) \in \mathcal{E} \\
    &(t_i, r_{i+1}, t_{i+1}) \in \mathcal{P}_\ell
\end{aligned}
\right\},
\end{equation}
where $i = 0, 1, \dots, K-1$, and $\mathcal{C}_v^{(\ell)}$ is the set of node sequences starting from $v$ and following the meta-path $\mathcal{P}_\ell$, $K$ is the walk length, $t_i$ and $r_{i+1}$ denote the node and relation types along the path, respectively. 
To learn unified structural embeddings across heterogeneous node and relation types, we employ skip-gram with negative sampling \cite{metapath2vec}, which brings together nodes that co-occur in meta-path contexts into a shared representation space. And the objective can be defined as:
\begin{align}
\mathcal{L}_{\text{struct}} = - & \frac{1}{L}  \sum_{\ell=1}^{L} 
    \ \ \ \mathbb{E}_{(v, u) \sim \mathcal{C}_v^{(\ell)}} 
        \left[ \log \sigma \left( \mathbf{z}_v^{\text{struct}^\top} \mathbf{z}^\text{struct}_u \right) \right] \nonumber \\
    &- \mathbb{E}_{(v, \tilde{u}) \sim P_n} 
        \left[ \log \left( 1 - \sigma \left( \mathbf{z}_v^{\text{struct}^\top} \mathbf{z}^\text{struct}_{\tilde{u}} \right) \right) \right],
\end{align}
where $P_n$ is a negative sampling distribution (e.g., uniform over all nodes), and all embeddings $\mathbf{z}^\text{struct}_v$ are learned jointly across all meta-paths. This loss encourages the embeddings to capture heterogeneous, high-order structural contexts in the graph. After obtaining the trained context structural embedding $\mathbf{z}^\text{struct}_v$, we freeze its parameters and concatenate the embedding with the original node attributes to unify the input of heterogeneous graph:
\begin{equation}
\tilde{\mathbf{x}}_v = \text{concat}(\mathbf{x}_v, \mathbf{z}_v^{\text{struct}}),
\end{equation}
where $\mathbf{x}_v$ denotes the original attributes of node $v$, and $\text{concat}(\cdot)$ denotes column-wise concatenation that appends the structural embedding as additional dimensions.
This contextual structural encoding strategy does not require type-specific transformation matrices or encoders for nodes and relations.
And we can encode both node and relation types into embedding $\mathbf{z}_v^{\text{struct}}$.

\textbf{Dimension-aware alignment.} While the contextual structural encoding provides unified node representation $\tilde{\mathbf{x}}_v$ for a single heterogeneous graph, its representation space can vary significantly across different graphs, which poses challenges for universal processing. 
To address this, we introduce a dimension-aware alignment module, inspired by the semantic basis learning strategy of \cite{FUG}.
We treat each attribute dimension as an independent semantic unit and learn a basis vector that reflects its latent global meaning.
For each input graph, we randomly sample $n_s$ nodes, and for the $i$-th attribute dimension, we aggregate its values as a column vector $\tilde{\mathbf{X}}_{:,i}^{\text{s}} \in \mathbb{R}^{n_s \times 1}$. 
We then apply a simple MultiLayer Perceptron (MLP) to encode it into a semantic basis vector $\mathbf{s}_i \in \mathbb{R}^k$:
\begin{equation}
\mathbf{s}_i = \text{MLP}(\tilde{\mathbf{X}}_{:,i}^{\text{s}}) =  (\tilde{\mathbf{X}}_{:,i}^{\text{s}})^\text{T} \mathbf{W} + \mathbf{b},
\end{equation}
where $\mathbf{W} \in \mathbb{R}^{n_s \times k}$ are learnable parameters, $\mathbf{s}_i \in \mathbb{R}^{1 \times k}$ is the basis vector for dimension $i$. 
Each node attribute $\tilde{\mathbf{x}}_v$ is then projected into a unified space $\mathbb{R}^k$ by aggregating the basis vectors weighted by the  dimension values:
\begin{equation}
    \mathbf{x}_v^{\text{unify}} = \sum_{i=1}^{d} \tilde{\mathbf{x}}_v[i] \cdot \mathbf{s}_i.
\end{equation}
We train the dimension encoder with a mean-centering loss, which encourages the mean of all basis vectors to stay close to the origin: 
\begin{equation}
    \mathcal{L}_{\mathrm{align}}(S) = \left\| \frac{1}{d} \sum_{i=1}^{d} \mathbf{s}_i \right\|_2^2.
\end{equation}
In this way ,we can prevent the learned basis vectors from local bias and ensures that all attribute dimensions are aligned in a consistent way.

\subsection{Heterogeneous Information Transfer}
Building on the unified input, it remains crucial to overcome the limited transferability, which is mainly reflected in two aspects: encoding and aggregation. 
First, in the encoding stage, existing methods typically require to design a distinct encoding function for the adjacency matrix generated by each meta-path, which ties the learned representations to a particular set of meta-paths and makes it difficult to transfer.
Second, in the aggregation stage, these approaches further depend on learning meta-path-specific attention weights to combine the representations from different encoding functions, which further limits transferability.

\textbf{Universal encoding.} To enable universal encoding for heterogeneous graphs, we investigate the nature of different meta-path views and find that most exhibit a high degree of homophily (i.e., nodes of the similar semantics tend to connect). The high degree of homophily makes it possible to use a single shared encoder $\text{GNN}_\text{shared}$ to encode the topological views generated by different meta-paths. Based on this, we need to design a objective function, which can guide the $\text{GNN}_\text{shared}$ to capture the common connectivity patterns of neighboring nodes under different meta-path views.
Inspired by the success of masked auto-encoder methods in effectively learning structural information  \cite{GraphMAE, HGMAE}, we apply a mask-and-reconstruct strategy for each meta-path view. 
Specifically, for each meta-path based adjacency matrix $\mathbf{A}^{\phi}$, we construct a binary mask $\mathbf{M}_{A}^{\phi} \sim \mathrm{Bernoulli}(p_e)$, where $p_e$ is the edge masking rate. The embedding of each views can be encoded as follows:
\begin{equation}
    \mathbf{Z}^\phi = \text{GNN}_\text{shared}(\tilde{\mathbf{A}}^{\phi},\mathbf{X}^{\text{unify}}), \tilde{\mathbf{A}}^{\phi} = \mathbf{M}_{A}^{\phi} \odot \mathbf{A}^{\phi}.
\end{equation}
We then apply a GNN decoder to obtain the decoded node embeddings $\hat{\mathbf{Z}}^\phi$ and get the reconstructed adjacency matrix by $\hat{\mathbf{A}}^\phi = \sigma(\hat{\mathbf{Z}}^\phi {\hat{\mathbf{Z}}^{\phi^T}})$. Then, a scaled cosine loss is used to train the model:
\begin{equation}
    \mathcal{L}^{\phi} = \frac{1}{|A^{\phi}|} \sum_{v \in V} \left( 1 - \frac{\langle A^{\phi}_v, \hat{A}^{\phi}_v \rangle}{\|A^{\phi}_v\| \cdot \|\hat{A}^{\phi}_v\|} \right)^{\gamma_1},
\end{equation}
where $\gamma_1$ is a scaling parameter. 

\textbf{Universal aggregation.} After obtaining the encoding loss $\mathcal{L}^{\phi}$ from different meta-paths, we need to design a universal aggregation mechanism. 
In conventional heterogeneous graph pre-training, a semantic-level attention vector $\mathbf{q}$ is introduced to learn attention weights $\beta^{\phi}$ for each meta-path: 
\begin{equation}
    c^{\phi} =  \mathbf{q}^\top \cdot \tanh(\mathbf{W} \cdot \mathbf{Z}^{\phi} + \mathbf{b}), \beta^{\phi} = \frac{\exp(c^{\phi})}{\sum_{\phi \in \Phi} \exp(c^{\phi})},
\end{equation}
where $\mathbf{W}$ is the weight matrix, $\mathbf{b}$ is the bias vector, $c^{\phi}$ is the contribution of meta-path $\phi$. However, such aggregation strategies are often tightly coupled to the training dataset and fail to generalize to new graphs or tasks. 
This is because the training objective leverages dataset-specific priors to shape the embedding distribution in the representation space, such as pulling semantically similar nodes closer or pushing dissimilar ones apart. As a result, the model tends to learn aggregation weights that are tailored to the semantic patterns of specific datasets, thereby limiting generalizability. To address this issue, we introduce an additional global scattering regularization loss inspired by \cite{sgrl}:
\begin{equation}
\mathcal{L}_{\text{scatter}} = -\frac{1}{|\mathcal{V}|} \sum_{v\in\mathcal{V}} \left\| \mathbf{z}_v - \bar{\mathbf{z}} \right\|_2^2,
\end{equation}
where $\bar{\mathbf{z}} = \frac{1}{|\mathcal{V}|}\sum_{v\in\mathcal{V}} \mathbf{z}_v$. Rather than guiding aggregation via dataset-specific signals, it encourages node embeddings to be pushed away from the global mean in the representation space, thereby increasing their discriminability. This regularization alleviates the reliance on specific aggregation functions and promotes more transferable node representations across heterogeneous graphs. And  we compute the final loss by the following equation:
\begin{equation}
\mathcal{L} = \lambda_1\mathcal{L}_\text{align} + \lambda_2 \sum_{\phi \in \Phi}\beta^\phi\mathcal{L}^\phi + \lambda_3\mathcal{L}_\text{scatter},
\end{equation}
where $\lambda_1,\lambda_2,\lambda_3$ are hyper-paramters to adjust the weight of loss. We optimize all trainable parameters through this loss.
\begin{table*}[ht]
\centering
\resizebox{1.0\textwidth}{!}{
\begin{tabular}{l|lcccccccc}
\toprule
\multicolumn{1}{l|}{\multirow{2}{*}{\textbf{Train}}} & \multicolumn{1}{l}{\multirow{2}{*}{\textbf{Method}}} & \multicolumn{2}{c|}{\textbf{ACM}} & \multicolumn{2}{c|}{\textbf{DBLP}} & \multicolumn{2}{c|}{\textbf{AMiner}} & \multicolumn{2}{c}{\textbf{Freebase}} \\
\cline{3-10}
& & \textbf{Ma-F1} & \textbf{Mi-F1}
  & \textbf{Ma-F1} & \textbf{Mi-F1}
  & \textbf{Ma-F1} & \textbf{Mi-F1}
  & \textbf{Ma-F1} & \textbf{Mi-F1} \\
\midrule 
\multirow{4}{*}{ACM}
& HeCo   & 80.22$\pm$2.45 & 79.71$\pm$3.23 & 76.76$\pm$0.44 & 77.97$\pm$0.41 & 24.48$\pm$1.31 & 51.18$\pm$6.21 & 31.22$\pm$1.33 & 40.67$\pm$0.76 \\ % 
& HGMAE  & 84.22$\pm$0.52 & 84.01$\pm$0.50 & 87.17$\pm$0.19 & 88.23$\pm$0.23 & 29.08$\pm$0.83 & 41.91$\pm$7.06 & 32.59$\pm$1.12 & 42.95$\pm$1.31 \\
& HERO   & 84.37$\pm$0.29 & 84.12$\pm$0.30 & 84.60$\pm$0.61 & 85.80$\pm$0.54 & 44.08$\pm$1.38 & 50.14$\pm$1.31 & 33.69$\pm$1.77 & 43.32$\pm$0.43 \\
& \textbf{MUG}   & \textbf{85.52$\pm$0.79} & \textbf{84.90$\pm$1.13} & \textbf{91.69$\pm$0.13} &\textbf{92.38$\pm$0.30} & \textbf{76.35$\pm$0.04} & \textbf{87.02$\pm$0.15} & \textbf{46.05$\pm$0.52} & \textbf{49.78$\pm$1.29} \\
\midrule
\multirow{4}{*}{DBLP}
& HeCo   & 82.93$\pm$0.80 & 83.43$\pm$0.70 & 90.11$\pm$0.33 & 90.73$\pm$0.31 & 26.89$\pm$1.09 & 35.23$\pm$5.45 & 35.58$\pm$0.94 & 39.53$\pm$2.11 \\
& HGMAE  & 83.99$\pm$0.47 & 83.44$\pm$0.82 & 89.97$\pm$0.30 & 90.89$\pm$0.27 & 33.95$\pm$0.37 & 45.42$\pm$4.30 & 33.81$\pm$0.97 & 41.16$\pm$2.00 \\
& HERO   & 84.92$\pm$0.46 & 84.68$\pm$0.48 & 87.47$\pm$0.36 & 88.57$\pm$0.34 & 51.66$\pm$1.30 & 61.04$\pm$2.93 & 32.63$\pm$2.62 & 42.99$\pm$0.65 \\
& \textbf{MUG}   & \textbf{85.81$\pm$0.50} & \textbf{85.06$\pm$0.54} & \textbf{90.67$\pm$0.28} & \textbf{91.40$\pm$0.33} & \textbf{73.44$\pm$1.05} & \textbf{84.96$\pm$0.42} & \textbf{50.48$\pm$0.97} & \textbf{55.22$\pm$0.89} \\
\midrule
\multirow{4}{*}{AMiner}
& HeCo   & 78.01$\pm$3.06 & 77.70$\pm$3.06 & 80.79$\pm$0.38 & 82.20$\pm$0.39 & 24.69$\pm$0.48 & 44.39$\pm$8.62 & 35.93$\pm$1.33 & 43.54$\pm$1.50 \\
& HGMAE  & 83.67$\pm$0.51 & 83.48$\pm$0.49 & 87.66$\pm$0.64 & 88.82$\pm$0.56 & 27.83$\pm$1.08 & 48.71$\pm$6.31 & 35.92$\pm$0.60 & 41.52$\pm$1.81 \\
& HERO   & 84.12$\pm$0.23 & 83.89$\pm$0.17 & 88.34$\pm$0.47 & 89.27$\pm$0.44 & 54.37$\pm$1.33 & 63.21$\pm$1.45 & 33.39$\pm$2.17 & 42.46$\pm$0.66 \\
& \textbf{MUG}   & \textbf{85.34$\pm$0.14} & \textbf{84.94$\pm$0.10} & \textbf{91.81$\pm$0.15} & \textbf{92.82$\pm$0.16} & \textbf{75.08$\pm$0.24} & \textbf{85.56$\pm$0.33} & \textbf{47.61$\pm$0.54} & \textbf{55.28$\pm$0.78} \\
\midrule
\multirow{4}{*}{Freebase}
& HeCo   & 77.03$\pm$0.89 & 76.30$\pm$1.44 & 82.37$\pm$0.47 & 83.26$\pm$0.47 & 29.82$\pm$0.82 & 34.51$\pm$4.34 & 42.34$\pm$1.68 & 47.92$\pm$2.32 \\
& HGMAE  & 84.78$\pm$0.54 & 84.90$\pm$0.51 & 83.97$\pm$0.94 & 84.65$\pm$1.25 & 24.16$\pm$0.94 & 47.26$\pm$6.44 & 33.17$\pm$0.84 & 41.07$\pm$1.75 \\
& HERO   & 84.65$\pm$0.36 & 84.35$\pm$0.37 & 84.29$\pm$0.57 & 85.84$\pm$0.62 & 48.26$\pm$1.13 & 58.19$\pm$1.17 & 31.25$\pm$1.73 & 42.14$\pm$0.98 \\
& \textbf{MUG}   & \textbf{85.21$\pm$1.26} & \textbf{85.22$\pm$1.05} & \textbf{91.79$\pm$0.28} & \textbf{92.24$\pm$0.17} & \textbf{78.10$\pm$1.35} & \textbf{87.94$\pm$0.56} & \textbf{52.33$\pm$0.26} & \textbf{57.50$\pm$2.41} \\
\bottomrule
\end{tabular}
}
\caption{Performance on cross-domain node classification with standard deviations. The best results are highlighted in bold.}
\label{tab:cross-domain-60}
\end{table*}
\section{Experiment}

To comprehensively evaluate the effectiveness of the proposed MUG, we conducted extensive experiments. In the following subsections, we first introduce the experimental setup. Subsequently, we give the performance analysis on cross-domain and few-shot node classification tasks and model analysis to understand the contribution of each component of the model.

\subsection{Experimental Setup}
\label{subsec:exp_setup}
\textbf{Datasets.} We follow the prior work \cite{HeCo} and evaluate the performance of MUG on four widely-used heterogeneous datasets: ACM \cite{ACM}, DBLP \cite{magnn}, AMiner \cite{Aminer} and Freebase \cite{freebase}.
Each dataset contains a target node type (in bold), and our downstream node classification task is conducted specifically on nodes of this type.

\begin{itemize}
    \item ACM contains three types of nodes: \textbf{papers}, authors, and subjects. It includes two types of relations (paper–author and paper–subject), and two meta-paths: PAP and PSP.
    \item DBLP contains four types of nodes: \textbf{authors}, papers, conferences, and terms. It includes three types of relations (author–paper, paper–conference, and paper–term), and three  meta-paths: APA, APCPA, and APTPA.
    \item AMiner contains three types of nodes: \textbf{papers}, authors, and references. It includes two types of relations (paper–author and paper–reference), and two meta-paths: PAP and PRP.
    \item Freebase contains four types of nodes: \textbf{movies}, actors, directors, and writers. It includes three types of relations (movie–actor, movie–director, and movie–writer), and three meta-paths: MAM, MDM, and MWM.
\end{itemize}

\textbf{Baselines.} As there is no existing work that focuses on universal pre-training for heterogeneous graphs, we compare MUG with several well-established self-supervised learning methods for heterogeneous graph representation. Specifically, we include HeCo \cite{HeCo}, a contrastive learning method that jointly captures local and high-order semantics from schema and meta-path views; HGMAE, a generative method that trains the model by reconstructing masked node attributes, meta-paths, and positional attributes; and HERO \cite{hero}, a recent method that jointly captures homophily and heterogeneity without relying on pre-defined meta-paths.

\textbf{Implementation Details.} To comprehensively evaluate the performance of MUG, we conducted experiments in two general application scenarios, i.e., cross-domain node classification and cross-domain few-shot node classification. 
For the cross-domain node classification, we train the model on one dataset and evaluate it on all datasets. During the downstream tasks, all model parameters are frozen. The training, validation, and test splits follow previous work \cite{HeCo}, with 60 nodes per class for training and 1,000 nodes each for validation and testing. 
To further evaluate performance with extremely limited labels, we conduct few-shot experiments with 1-shot, 3-shot, and 5-shot settings, where each training class label in ths test dataset is provided with only 1, 3, or 5, respectively.

\subsection{Performance Analysis}
\label{subsec:exp_performance_analysis}
\textbf{Cross-Domain node classification.} In this scenario, we follow the evaluation protocol in \cite{HeCo}, reporting the test performance when validation set achieves the best result. All experiments are repeated 50 times, and the mean and standard deviation are reported. We adopt both Macro-F1 and Micro-F1 scores as evaluation metrics. As shown in Table \ref{tab:cross-domain-60}, MUG consistently outperforms all baseline models across all datasets, demonstrating its superior cross-domain transfer capability and robust generalization.
\begin{table*}[ht]
\centering
\resizebox{1.0\textwidth}{!}{
\begin{tabular}{l|lcccccccc}
\toprule
\multicolumn{1}{l|}{\multirow{2}{*}{\textbf{Shot-num}}} & \multicolumn{1}{l}{\multirow{2}{*}{\textbf{Method}}} & \multicolumn{2}{c|}{\textbf{ACM}} & \multicolumn{2}{c|}{\textbf{DBLP}} & \multicolumn{2}{c|}{\textbf{Aminer}} & \multicolumn{2}{c}{\textbf{Freebase}} \\
\cline{3-10}
& & \textbf{Ma-F1} & \textbf{Mi-F1}
  & \textbf{Ma-F1} & \textbf{Mi-F1}
  & \textbf{Ma-F1} & \textbf{Mi-F1}
  & \textbf{Ma-F1} & \textbf{Mi-F1} \\
\midrule 
\multirow{4}{*}{1-shot}
& HeCo   & 59.36$\pm$12.32  & 63.41$\pm$9.52  & 45.74$\pm$10.32  & 49.94$\pm$9.53  
& 23.85$\pm$2.61   & 39.69$\pm$10.43 & 30.70$\pm$3.08 & 39.58$\pm$3.68 \\
& HGMAE & 73.17$\pm$5.04 & 75.48$\pm$4.96  & 61.46$\pm$5.68 & 65.94$\pm$5.54  
& 20.65$\pm$4.46  & 36.88$\pm$7.33  & 30.65$\pm$3.87  & 38.20$\pm$4.37 \\
& HERO   & 51.39$\pm$1.99 & 57.37$\pm$1.31 & 40.49$\pm$4.12 & 41.87$\pm$4.88 & 44.18$\pm$0.78 & 50.14$\pm$1.31 & 32.20$\pm$2.21 & 42.56$\pm$0.97 \\
& \textbf{MUG}   & \textbf{79.49$\pm$4.26} & \textbf{79.54$\pm$3.93} & \textbf{84.24$\pm$5.64} &\textbf{85.76$\pm$4.95} & \textbf{49.12$\pm$6.16} & \textbf{72.10$\pm$4.68} & \textbf{33.24$\pm$2.63} & \textbf{42.20$\pm$1.96} \\
\midrule
\multirow{4}{*}{3-shot}
& HeCo   & 73.53$\pm$8.71  & 74.77$\pm$7.29  
& 64.49$\pm$6.69  & 66.09$\pm$6.50  
& 24.93$\pm$1.92  & 40.56$\pm$8.13
 & 33.15$\pm$2.66 & 40.12$\pm$2.39 \\
& HGMAE  & 79.42$\pm$7.82  & 79.58$\pm$7.53  & 71.39$\pm$8.10 & 73.88$\pm$7.61  
& 23.73$\pm$3.80  & 41.71$\pm$8.45  & 32.47$\pm$2.75  & 40.74$\pm$3.10\\
& HERO   & 63.78$\pm$2.25 & 63.74$\pm$2.23 & 61.24$\pm$0.67 & 61.50$\pm$0.94 & 47.38$\pm$0.67 & 53.17$\pm$1.26 & 34.17$\pm$1.32 & 42.63$\pm$1.49 \\
& \textbf{MUG}   & \textbf{84.39$\pm$2.48} & \textbf{83.74$\pm$2.87} & \textbf{90.56$\pm$0.66} & \textbf{91.62$\pm$0.49} & \textbf{66.80$\pm$2.85} & \textbf{81.92$\pm$2.64} & \textbf{35.01$\pm$1.14} & \textbf{44.48$\pm$1.98} \\
\midrule
\multirow{4}{*}{5-shot}
& HeCo   & 77.58$\pm$4.92  & 78.08$\pm$4.19  
& 68.75$\pm$4.53  & 70.05$\pm$4.71  
& 25.47$\pm$2.12  & 40.86$\pm$7.60
 & 33.50$\pm$2.08 & 40.55$\pm$1.94 \\
& HGMAE  & 81.68$\pm$4.72  & 81.26$\pm$5.32  & 79.03$\pm$6.92  & 80.39$\pm$7.00  
& 24.84$\pm$3.10  & 41.96$\pm$9.75  & 33.45$\pm$2.20  & 40.46$\pm$3.46 \\
& HERO   & 74.08$\pm$1.14 & 73.76$\pm$1.14 & 63.31$\pm$0.80 & 63.42$\pm$0.84 & 44.18$\pm$0.78 &50.14$\pm$1.31 & 32.20$\pm$2.21 &42.56$\pm$0.97 \\
& \textbf{MUG}   & \textbf{83.83$\pm$1.10} & \textbf{82.64$\pm$2.16} & \textbf{90.76$\pm$0.37} & \textbf{91.80$\pm$0.38} & \textbf{68.30$\pm$1.56} & \textbf{83.76$\pm$1.60} & \textbf{39.36$\pm$4.08} & \textbf{45.82$\pm$1.93} \\
\bottomrule
\end{tabular}
}
\caption{Performance on few-shot node classification with standard deviations. The best results are highlighted in bold.}
\label{tab:cross-domain-few-shot}
\end{table*}
For all three baseline methods, we apply SVD\cite{SVD} to unify the input space across datasets. However, this approach inevitably leads to the loss of rich semantic information present in heterogeneous data. In contrast, our contextual structure encoder explicitly encodes both node types and relation types, and utilizes a dimension encoder to achieve alignment in the representation space. This design enables MUG to preserve and leverage the diverse semantics inherent in heterogeneous graphs, supporting more effective transfer across domains. Additionally, we observe that the performance gap between MUG and the baseline methods is particularly pronounced on the AMiner dataset. This is mainly because node attributes in AMiner are represented as one-hot vectors, making them especially susceptible to information loss under dimension reduction. This observation further highlights the importance of our approach in unifying heterogeneous inputs and maintaining the integrity of original semantic information.

\begin{figure}[H]
\includegraphics[width=1.0\linewidth]{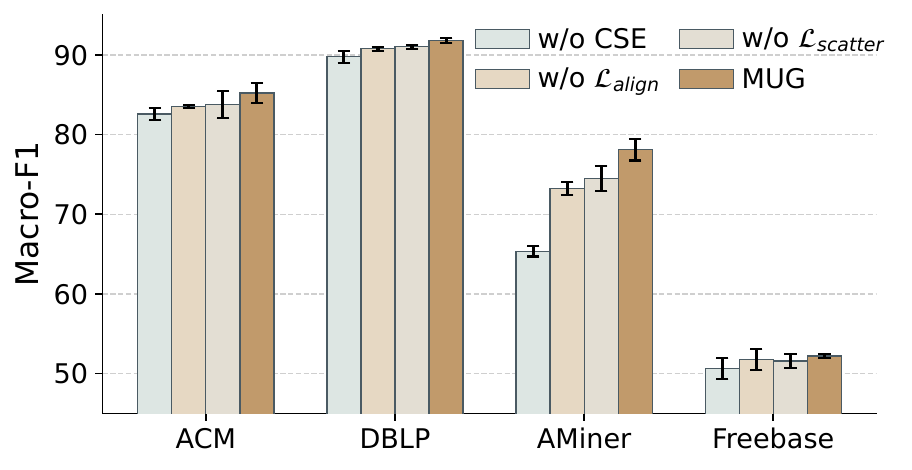}
    \caption{Ablation results for cross-domain node classification, trained on Freebase and evaluated on four datasets.}
    \label{fig:ab_study}
\end{figure}

\textbf{Cross-Domain few-shot node classification.} For the cross-domain few-shot node classification experiments, we follow the same evaluation protocol as described above. In this setting, we trained MUG on the ACM dataset and directly evaluated it on the all datasets using only 1, 3, or 5  labeled train samples per class. As shown in Table \ref{tab:cross-domain-few-shot}, MUG achieves the best performance among all methods under all few-shot settings. Notably, in the 5-shot scenario, the performance of MUG approaches the results reported in Table \ref{tab:cross-domain-60} for the cross-domain node classification setting.
This result demonstrates the strong generalization and transfer capability of our method, even when the amount of available labeled data is limited.
The consistently high performance of MUG in few-shot scenarios highlights the effectiveness of its unified representation space and its ability to capture transferable heterogeneous information during pre-training. 

\subsection{Model Analysis}
\label{subsec:exp_model_analysis}
Figure \ref{fig:ab_study} presents the results of ablation study under cross-domain node classification, where the model is trained on the Freebase dataset and evaluated on all four datasets. Here, CSE refers to the Contextual Structural Encoding component. Removing CSE consistently leads to the largest performance degradation, especially on cross-domain datasets such as AMiner, indicating that contextual structural encoding is crucial for capturing universal heterogeneous semantics beyond node attributes.
Moreover, removing the alignment loss $\mathcal{L}_{align}$ or the scattering loss $\mathcal{L}_{scatter}$ also results in noticeable performance drops.
This suggests that both losses play key roles in improving cross-domain generalization by mitigating domain-specific biases and promoting more transferable representations. Overall, the full model (MUG) achieves the best performance across all datasets.

\section{Conclusion}
In this work, we conduct an initial exploration into extending universal graph pre-training to heterogeneous graphs. We identify two key challenges: (i) the difficulty of unifying input due to the complex heterogeneity (i.e., node types,relation types and node attributes); and (ii) the limited transferability of heterogeneous information due to dataset-specific encoding and aggregation functions. To address these challenges, we propose MUG, a novel approach designed to learn transferable representations across diverse heterogeneous graphs. By leveraging a meta-path–based contextual encoding and a dimension-aware encoder, MUG can construct a unified representation space. Furthermore, MUG introduces a universal encoding and aggregation functions, facilitating effective information transfer while avoiding overfitting to specific datasets. Extensive experiments on cross-domain and few-shot settings demonstrate the strong performance of MUG.

\section{Acknowledgments}
This work was supported by the National Natural Science Foundation of China (No. 62422210, No. 62276187, No. 62372323), the National Key Research and Development Program of China (No. 2023YFC3304503),  Research Grants Council of Hong Kong through the Theme-based Strategic Topics Grant Scheme (STG1/M-501/23-N), the Hong Kong Global STEM Professor Scheme, and the Hong Kong Jockey Club Charity Trust.

\bibliography{aaai2026}

\end{document}